\documentclass[lettersize,journal]{IEEEtran}
\usepackage{amsmath,amsfonts}
\usepackage{algorithmic}
\usepackage{algorithm}
\usepackage{array}
\usepackage[caption=false,font=normalsize,labelfont=sf,textfont=sf]{subfig}
\usepackage{textcomp}
\usepackage{stfloats}
\usepackage{url}
\usepackage{verbatim}
\usepackage{graphicx}
\usepackage{cite}

\usepackage{makecell}
\usepackage{multirow}
\usepackage[table,xcdraw]{xcolor}
\usepackage[normalem]{ulem}
\useunder{\uline}{\ul}{}
\usepackage[colorlinks,linkcolor=red]{hyperref}

\begin{document}

\title{Voxel-Mesh Hybrid Representation For Real-Time View Synthesis By Meshing Density Field}

\markboth{Journal of \LaTeX\ Class Files,~Vol.~xx, No.~x, Aril~2024}%
{Shell \MakeLowercase{\textit{et al.}}: A Sample Article Using IEEEtran.cls for IEEE Journals}

\author{Chenhao Zhang, Yongyang Zhou and Lei Zhang*,~\IEEEmembership{Member,~IEEE,}
\thanks{Chenhao Zhang, Yongyang Zhou and Lei Zhang are with the School of Computer Science, Beijing Institute of Technology, Beijing 100081, China.}
\thanks{* Corresponding author: leizhang@bit.edu.cn}
}

\maketitle

\begin{abstract}
The neural radiance fields (NeRF) have emerged as a prominent methodology for synthesizing realistic images of novel views. While neural radiance representations based on voxels or mesh individually offer distinct advantages, excelling in either rendering quality or speed, each has limitations in the other aspect. In response, we propose a hybrid representation named \emph{Vosh}, seamlessly combining both voxel and mesh components in hybrid rendering for view synthesis. Vosh is meticulously crafted by optimizing the voxel grid based on neural rendering, strategically meshing a portion of the volumetric density field to surface.
Therefore, it excels in fast rendering scenes with simple geometry and textures through its mesh component, while simultaneously enabling high-quality rendering in intricate regions by leveraging voxel component. The flexibility of Vosh is showcased through the ability to adjust hybrid ratios, providing users the ability to control the balance between rendering quality and speed based on flexible usage. Experimental results demonstrate that our method achieves commendable trade-off between rendering quality and speed, and notably has real-time performance on mobile devices. The interactive web demo and code are available at \url{https://zyyzyy06.github.io/Vosh}.
\end{abstract}


\begin{figure*}[htbp]
  \includegraphics[width=\textwidth]{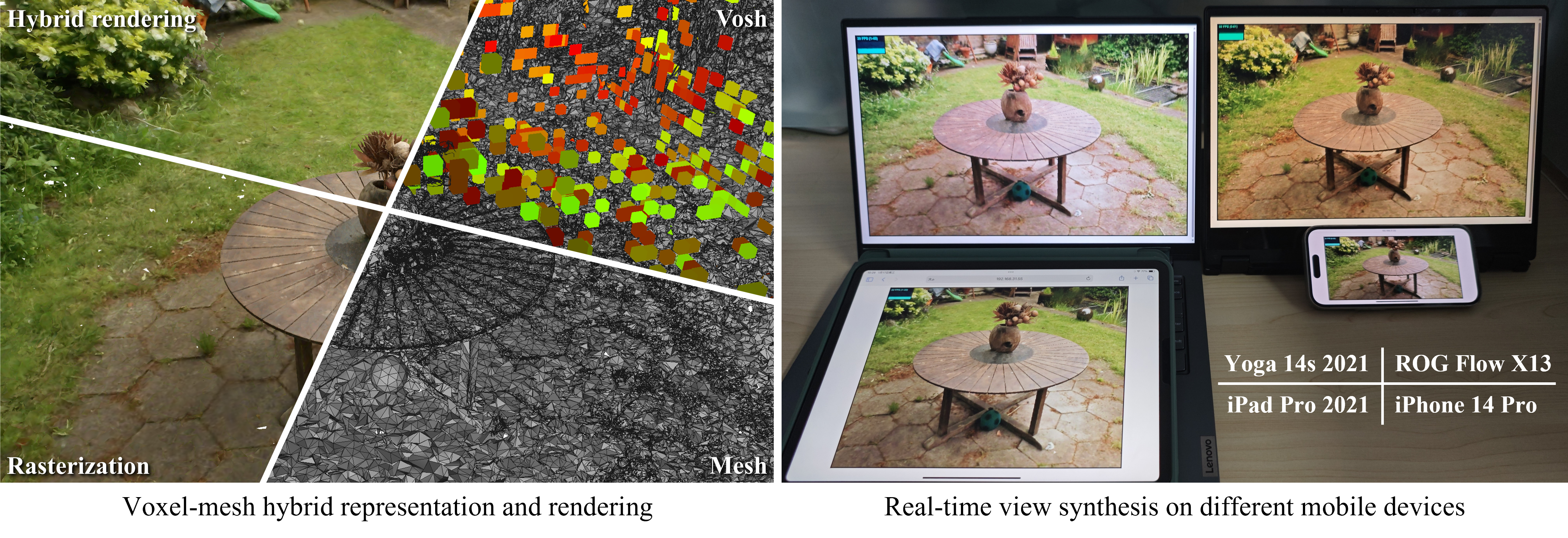}
  \caption{\textbf{Left:} The hybrid representation \emph{Vosh}, combines both voxel component (highlighted with pseudo colors) and mesh component (indicated by gray triangles) in hybrid rendering. \textbf{Right:} The proposed method facilitates real-time view synthesis  on various mobile devices, e.g., laptops and mobile phones.}
  \label{fig:teaser}
\end{figure*}

\section{Introduction}
The novel view synthesis based on the Neural Radiance Fields (NeRF) leverages neural networks to model three-dimensional scenes and generates images from specified viewpoints~\cite{nerf}. In general, NeRF-based approaches demonstrate proficiency in producing high-quality rendering for novel views with sparse inputs, effectively handling materials with specular and translucent properties. This versatility positions them for diverse applications in autonomous navigation~\cite{navigation}, virtual reality/augmented reality~\cite{vr}, and beyond~\cite{ide-3d,nerffacelighting}. Despite these merits, current methods face challenges in achieving a balance between high-quality rendering and real-time rendering, particularly when deployed on mobile devices. 

The rendering speed of neural radiance fields is significantly influenced by the chosen representation, with notable implications for sampling requirements and query costs~\cite{plenoctrees}. Typically, the original NeRF uses deep multilayer perceptrons (MLPs) to model neural radiance fields, relying on dense sampling for querying to approximate  integrals in volume rendering. Unfortunately, this approach typically causes excessively long rendering times. A substantial amount of effort has explored alternatives such as voxel representations~\cite{dvgo, snerg, nsvf, plenoxels, instantngp, kilonerf} and mesh representations~\cite{mobilenerf, nerf2mesh, nerfmeshing, bakedsdf}, aiming to exclude irrelevant regions for computation. These methods reduce the number of sampling points and employ smaller MLPs or no MLPs to expedite rendering. However, even though approaches based on voxels can achieve high-quality rendering in complex scenes, the dense sampling in non-empty regions is prohibitively expensive. 3DGS~\cite{3dgs}, with its Gaussian representation and Splatting rendering method, also reduces the number of sampled points in rays, achieving real-time rendering of high-resolution scenes. But its high memory usage and platform-related sorting strategies hinder its widespread applications, especially on mobile devices.

The methods based on mesh representation have demonstrated significantly accelerated rendering speeds, enabling real-time rendering on common mobile devices. However, the challenge lies in the reconstruction of scene-level mesh structure, particularly in areas with intricate geometry and textures (see Figure~\ref{fig:teaser}), such as thin structures or undetermined regions like backgrounds~\cite{merf}. It is observed that meshes can match the rendering quality akin to voxels in regions with simple geometry and textures, as well as offering highly efficient rendering. Therefore, a promising solution involves employing hybrid representations for reconstruction towards both high-quality rendering and fast speed. 

Existing works on hybrid representations~\cite{adaptiveshells,vmesh}, focus primarily on object or proximal region reconstruction, thus abandoning the reconstruction ability of volume representation for complex scenes, such as background regions lacking observation in unbounded outdoor environments. VMesh~\cite{vmesh} mainly relies on the signed distance function  for the scene reconstruction. However, it may result in low-quality reconstruction and is space-consuming for complex background areas of unbounded scenes. We utilize implicit volumetric density field, which can recover more comprehensive information and be applied in both bounded and unbounded scenes. By partially meshing the density field, a more expressive mesh representation can be obtained. Then we leverage the strengths of voxels and mesh, to design a hybrid representation towards a better balance between quality and speed, especially enabling high-quality rendering on mobile devices (see Figure~\ref{fig:teaser}).

In summary, the main contribution of this paper is a new hybrid representation namely \emph{Vosh} (VOxel-meSH), by combing voxels and mesh for NeRF-based real-time view synthesis. It also allows for a dynamic control of the balance between voxel and mesh components. Such flexibility results in a versatile spectrum of NeRF representations, catering to the varied computational capabilities that facilitate real-time view synthesis across different devices. Experimental results exhibit the superior performance of our method compared to state-of-the-art methods. 

\section{Related Work}

There are numerous published approaches to view synthesis and NeRF, and we refer interested readers to~\cite{advances, nerfreview, beyondpixels} for a comprehensive overview. In this context, we focus on the real-time view synthesis and NeRF representations most related to our method.

\subsection{Real-time view synthesis}

NeRF~\cite{nerf} as a well-known technology for view synthesis, has been the focus of numerous endeavors aimed at accelerating its rendering speed while maintaining high quality of rendered images. SNeRG~\cite{snerg} achieves fast rendering without CUDA, using sparse voxel representations and a compact neural network for deferred rendering. Plenoxels~\cite{plenoxels} and PlenOctrees~\cite{plenoctrees} expedite ray marching with sparse voxel representation, refining voxels near object surfaces. MERF~\cite{merf} enhances rendering speed and spatial occupancy by replacing pure voxel grids with a low-resolution voxel grid and a high-resolution triplane projection representation~\cite{eg3d}. These methods mostly excel on high-end GPUs, whereas real-time rendering on mobile devices still remains a challenge.

MobileNeRF~\cite{mobilenerf} achieves fast rendering by constraining NeRF to triangles and using rasterization. NeRF2Mesh~\cite{nerf2mesh} uses NeRF to initialize geometric structures, iteratively refining mesh attributes to improve rendering quality. BakedSDF~\cite{bakedsdf} optimizes a blended neural signed distance function representation based on VolSDF~\cite{volsdf}, enabling real-time rendering of complex scenes on a high-quality mesh. In order to represent complex backgrounds, BakedSDF uses a contracted space containing Eikonal loss which might have negative impact on the quality of the foreground as claimed in~\cite{hybridnerf}. While mesh-based methods offer fast rendering, they suffer quality loss compared to volume-based methods, especially for unbounded scenes with complex backgrounds.

3D Gaussian Splatting~\cite{3dgs} creates a 3D Gaussian representation around each point starting from a point cloud model, utilizing the Splatting method for real-time rendering of high-resolution scenes. To achieve real-time training and rendering, 3DGS relies on CUDA-specific APIs. So original 3DGS is difficult to render on most mobile devices. While recent re-implementations seem to render 3DGS models on commodity hardware, they rely on approximations to sort order and view-dependency whose impact on quality and speed has not yet been evaluated~\cite{smerf}. In contrast, our hybrid representation combines the strengths of voxel and mesh representations, allowing high-quality rendering of complex scenes on mobile devices. 

\textcolor{black}{Recently, 3DGS-based compression methods have emerged to effectively reduce the number of Gaussians. Eagles~\cite{Eagles} proposes to use quantized embeddings and a coarse-to-fine training strategy to significantly reduce memory storage overhead. F-3DGS~\cite{F-3DGS} represents dense clusters of Gaussians with significantly fewer Gaussians through efficient factorization. While these methods have successfully enhanced 3DGS in terms of storage efficiency and rendering speed, they still require as much or even more VRAM in the rendering compared to the original 3DGS as reported in these papers, which are still not suitable for the use on mobile devices. }

\subsection{Hybrid scene representation}

Combining explicit mesh representation with implicit fields paves a new way to NeRF-based processing, such as NeRF editing and rendering acceleration. NeuMesh~\cite{neumesh} encodes a neural implicit field with disentangled geometry and texture on mesh vertices, enabling a range of neural rendering-based editing operations. EyeNeRF~\cite{eyenerf} integrates an explicit surface with an implicit volume representation to model the human eye. Mesh-Aware-RF~\cite{Mesh-Aware-RF} embeds a known polygonal mesh into NeRF, employing ray tracing and ray marching alternately to update radiance along rays.

The hybrid representation like Adaptive Shells~\cite{adaptiveshells} and VMesh~\cite{vmesh} has been used in view synthesis. Adaptive Shells confine volume rendering to a narrow band around the object for effective NeRF rendering. VMesh, closely related to our work, uses a textured mesh and auxiliary sparse voxels but differs by employing the signed distance function (SDF) and volume density field to simultaneously learn the bounded scenes or objects. Both methods use NeuS~\cite{neus} as the implementation for SDF. NeuS learns neural implicit surfaces through volume rendering. For the unbounded scenes with backgrounds, NeuS needs to use NeRF for background modeling that is difficult to represent by itself, which is also a hybrid representation.

\textcolor{black}{It should be noted that VMesh most closely resembling our method in this paper, relies on SDF with density field to capture scene information, while our method only uses the density field. Specifically, when using SDF to reconstruct complex areas like cluttered backgrounds in unbounded scenes, it might lead to low-quality reconstruction and significant space occupation}~\cite{bakedsdf}\textcolor{black}{. Due to this limitation, VMesh is primarily suited for single objects or bounded scenes, whereas our density-field-based method effectively manages a wide range of scenarios, including unbounded outdoor scenes. Besides, VMesh requires manual tuning of multiple loss weights for each scene to achieve optimal surface-volume separation as indicated in the paper~\cite{vmesh}. Without these adjustments, only a single representation may be optimized. In contrast, our  method based on the density field does not rely on scene-specific hyper-parameters, enabling the high-quality hybrid representation for various scenes.}

\begin{figure*}[htbp]
  \includegraphics[width=\textwidth]{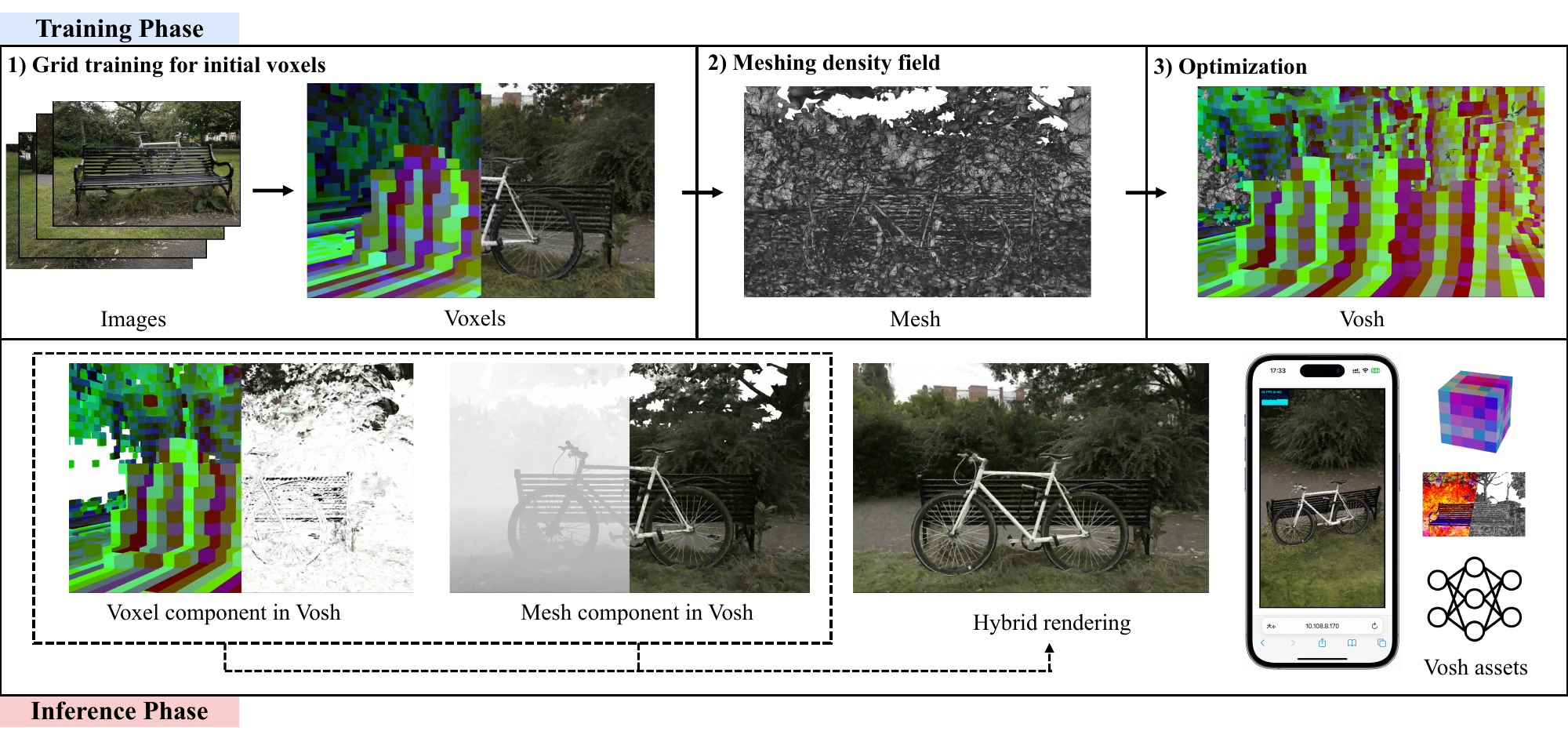}
  \caption{\textbf{An overview of the proposed methodology}. The training phase starts from grid training for obtaining initial voxels. Then, a portion of initial volumetric density field are meshed. Subsequently, the combination of voxels and mesh are optimized through hybrid rendering and voxel pruning to obtain the final hybrid representation Vosh. The inference phase realizes real-time hybrid rendering with Vosh even on mobile phones.}
  \label{fig:pipeline}
\end{figure*}

\section{Method Overview}
 
We start with an overview of our methodology to establish a context for the core algorithms described in the next section. As depicted  in Figure~\ref{fig:pipeline}, our method contains two key phases: the training phase for constructing a hybrid representation and the  inference phase for view synthesis rendering. In the training phase, a set of multi-view images is employed to train a voxel-mesh hybrid representation, and the associated assets are exported. In the inference phase, our method achieves real-time rendering based on the baked hybrid representation assets.

Concretely, the \textbf{training phase} consists of three stages: grid training, meshing density field and optimization. In the grid training stage, we train a voxel grid model through volume rendering. Then, we determine the region of meshing in the density field based on the difference in rendering quality between the voxel grid and the potential surfaces it contains. This converts the initial voxel grid into a combination of voxel and mesh components. Finally,  the voxels and mesh are further optimized through a training process involving hybrid rendering and voxel pruning to obtain the desired hybrid representation. Once training is complete, we export assets containing sparse voxels and textured meshes. 
Subsequently, in the \textbf{inference phase}, we implement the rendering process of Vosh based on WebGL, enabling real-time view synthesis on mobile devices.
Next, we elaborate the algorithm details in the training and inference phases.

\section{Hybrid Representation construction}

The construction of a hybrid representation in the training phase commences with the initialization of a volume comprising solely of voxels. Then, by calculating the difference in rendering quality between the volume and its underlying surface, the density field corresponding to regions with similar rendering quality are meshed to replace the volumetric density values corresponding to the same positions.

\subsection{Grid training for initial voxels}
Initially, the entire scene is reconstructed using a high-resolution voxel grid. It retains sufficient geometric and appearance information essential for subsequent mesh extraction and hybrid representation. Here, we employ SNeRG++ (an improved version of SNeRG~\cite{snerg} in MERF~\cite{merf}) as the benchmark for our voxel grid, due to its capability to effectively model unbounded scenes and generate real-time rendering assets. 
    
Concretely, SNeRG++ is built upon the volume rendering principles of NeRF, which involves sampling along the direction of a ray originating from the ray origin $\mathbf{o}$ along the direction $\mathbf{d}$ to obtain sample points $\mathrm{\mathbf{x}}_i=\textbf{o}+t_i\mathrm{\mathbf{d}}$. Based on the spatial position of these sample points, it uses the neural radiance field to calculate volumetric density {$\sigma_i$}, resulting in rendering weights {$w_i$}:
\begin{equation}
    \label{eqn:1}
    w_i=\alpha_iT_i,\;\;\;\;\;\;\;\;T_i=\prod^{i-1}_{j=1}(1-\alpha_j),\;\;\;\;\;\;\;\;\alpha_i=1-e^{-\sigma_i\delta_i},
\end{equation}
where $T_i$, $\alpha_i$ and $\delta_i$ represent the transmittance, opacity and spacing of sampling point $i$, respectively. SNeRG++ adopts the concept of deferred rendering, decomposing the appearance information of the neural radiance field into diffuse color {$\mathrm{\mathbf{c}}_{d,i}$} and view-dependent features {$\mathbf{f}_i$}. By cumulatively sampling the appearance information along the ray, we can obtain the diffuse color $\mathbf{C}_d$ and view-dependent features $\mathbf{F}$ of the ray as
\begin{equation}
    \label{eqn:2}
\mathrm{\mathbf{C}}_d=\sum_iw_i\mathrm{\mathbf{c}}_{d,i},\;\;\;\;\;\;\;\;\mathrm{\mathbf{F}}=\sum_iw_i\mathrm{\mathbf{f}}_i.
\end{equation}
Finally, the view-dependent color is calculated by inputting the diffuse color $\mathrm{\mathbf{C}}_d$, view-dependent features $\mathbf{F}$, and the ray direction $d$ into a miniature MLP. The result is then added to the diffuse color $\mathrm{\mathbf{C}}_d$ to obtain the final color $\mathbf{C}$ of the ray.

Besides, in order to enhance the modeling of regions situated far from the center within unbounded scenes and facilitate efficient ray marching calculations, SNeRG++ incorporates a contraction function defined on piecewise projection. 

\subsection{Meshing density field}
\textcolor{black}{The motivation for  meshing density field is to convert as many voxels as possible to mesh with minimal quality loss. Usually, there is a significant quality gap between voxel and mesh representation, but the distribution corresponding to the gap is unknown. We identify spatial regions where the volume and its implicit surface offer comparable rendering quality by measuring the quality difference between the surface after differentiable refinement and initial voxels, as well as meshing the density field within these regions. In this way, we can achieve speed gain with minimal quality loss.}

\textbf{Differentiable surface refinement}.\quad 
Based on the above high-resolution voxel grid, we employ Marching Cubes~\cite{marchingcubes} for the density field of initial voxel grid to extract an explicit mesh for evaluating the quality gap between voxel and mesh representation. However, surfaces meshed directly from volumetric density field often exhibit roughness that have to be improved for enhancing the rendering quality. Inspired by NeRF2Mesh~\cite{nerf2mesh},  we adopt an ad-hoc scheme involving differentiable rendering and surface operations to refine the explicit mesh firstly.

Concretely, we perform differentiable rasterization operations on the mesh to obtain intersections between rays and surfaces. Appearance information is obtained by querying the voxel representation based on the corresponding spatial positions. Similar to NeRF2Mesh~\cite{nerf2mesh}, we employ nvdiffrast~\cite{nvdiffrast} for differentiable rendering to optimize the appearance and the trainable offset of vertices, while simultaneously tracking the rendering error corresponding to each face for surface optimization. 

In addition to rendering errors, we also calculate the normal angle between the adjacent faces of each face for performing mesh refinement and decimation, thereby improving the representation ability of the mesh and reducing memory usage. Specifically, for coarse mesh extracted based on initial voxel grid, we use differentiable rendering to learn offsets for each vertex of the mesh. Meanwhile, we calculate the rendering error $E_{render}$ and the normal angle between the adjacent faces $A_{normal}$ for each face. We set the rendering error threshold $e_{sub}$ and angle threshold $a_{sub}$ for face subdivision based on percentiles:
\begin{equation}
    e_{sub}=percentile(E_{render}, 90),
\end{equation}
\begin{equation}
    a_{sub}=percentile(A_{normal}, 95).
\end{equation}

Similarly, we set the rendering error threshold $e_ {dec}$ and angle change threshold $a_{dec}$ for face simplification:
\begin{equation}
    e_{dec}=percentile(E_{render}, 50),
\end{equation}
\begin{equation}
    a_{dec}=percentile(A_{normal}, 10).
\end{equation}

For faces with $E_{render}>e_{sub}$ or $A_{normal}>a_{sub}$, we perform mid-point subdivision~\cite{meshlab} to increase face density and enhance representation ability. For faces with $E_{render}<e_{dec}$ or $A_{normal}<a_{dec}$, we perform faces simplification~\cite{decimation} to improve rendering efficiency.

In contrast to NeRF2Mesh~\cite{nerf2mesh}, which solely relies on rendering errors for operations, we additionally utilize the angle of local normal of faces as an optimization criterion. This enhances face density in regions with complex geometry and reduces storage overhead for faces in areas with simple geometry.

\begin{figure}[tbp]
  \includegraphics[width=\linewidth]{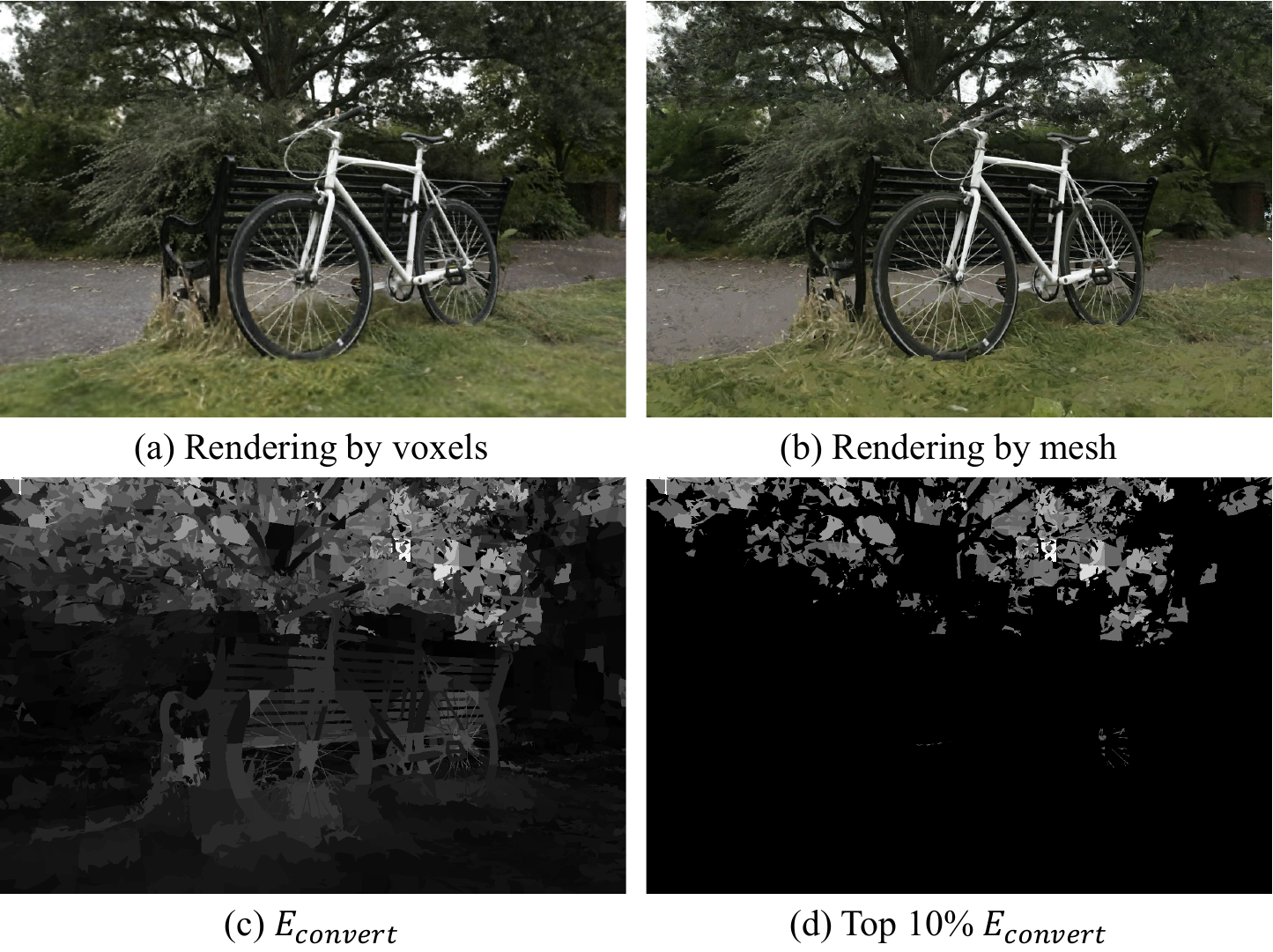}
  \caption{Visualization of conversion loss $E_{convert}$ based on rendering quality. (a) and (b) are the rendering images based on initial voxels and mesh with differentiable surface refinement, respectively. (c) is the gray-scale image of conversion loss $E_{convert}$ calculated based on the rendering quality of both representations. (d) is to filter out the top $10\%$ $E_{convert}$ in the space.}
  \label{fig:vis_error_convert}
\end{figure}

\textbf{Meshing based on rendering quality}.\quad
We can obtain the conversion error field $E_{convert}$ of different representations in the space based on differentiable rendering. Specifically, for voxel representation, we allocate rendering errors to corresponding positions based on the rendering weights of different sampling points. For mesh representation, the rendering errors at the corresponding position is recorded at the intersection of the surface and rays. We use grids to accumulate rendering errors for different representations and calculate the average rendering errors $E_{voxel}$ for voxel representation and $E_{mesh}$ for mesh representation. We calculate the L1 distance $E_{convert}=E_{mesh}-E_{voxel}$ of rendering errors in regions where neither is zero as the conversion errors, as shown in Figure~\ref{fig:vis_error_convert}.

It is observed in Figure~\ref{fig:vis_error_convert} that the main distribution of $E_{convert}$ is in the background area far from the center of cameras. This is probably because 1) the reconstruction quality of the central area which is fully observed by cameras is higher enough in mesh representation; 2) contraction function leads to significant loss of information in background voxels. So the meshing operation is focused on the background area that lacks sufficient observation and is affected by the contraction function. For the scenes based on contraction function in $[-2,2]$, we define the central region as $[-0.5, 0.5]$ and the outside as the background region. Then, we count voxels with background area and sort them, and also set the threshold in percentiles as $e_{remove} = percentile(E_{convert}, p_{remove})$. 
When $p_{remove}$ is set to 0.9, the voxels with their $E_{convert}$ values greater than the threshold $e_{remove}$ is shown in Figure~\ref{fig:vis_error_convert}(d).
Finally, we remove the faces converted from voxels located in the background area with $E_{convert} > e_{remove}$. The remaining mesh is the result of meshing the regions with small gap in rendering quality in the density field.

Following the rendering quality based meshing process in density field, we have obtained an initial hybrid representation that includes both voxels and mesh. Nevertheless, considering significant potential for optimization in rendering quality and volume occupancy for both representations, we are committed to further refining the hybrid representation through the next optimization.

\begin{figure}[tbp]
  \includegraphics[width=\linewidth]{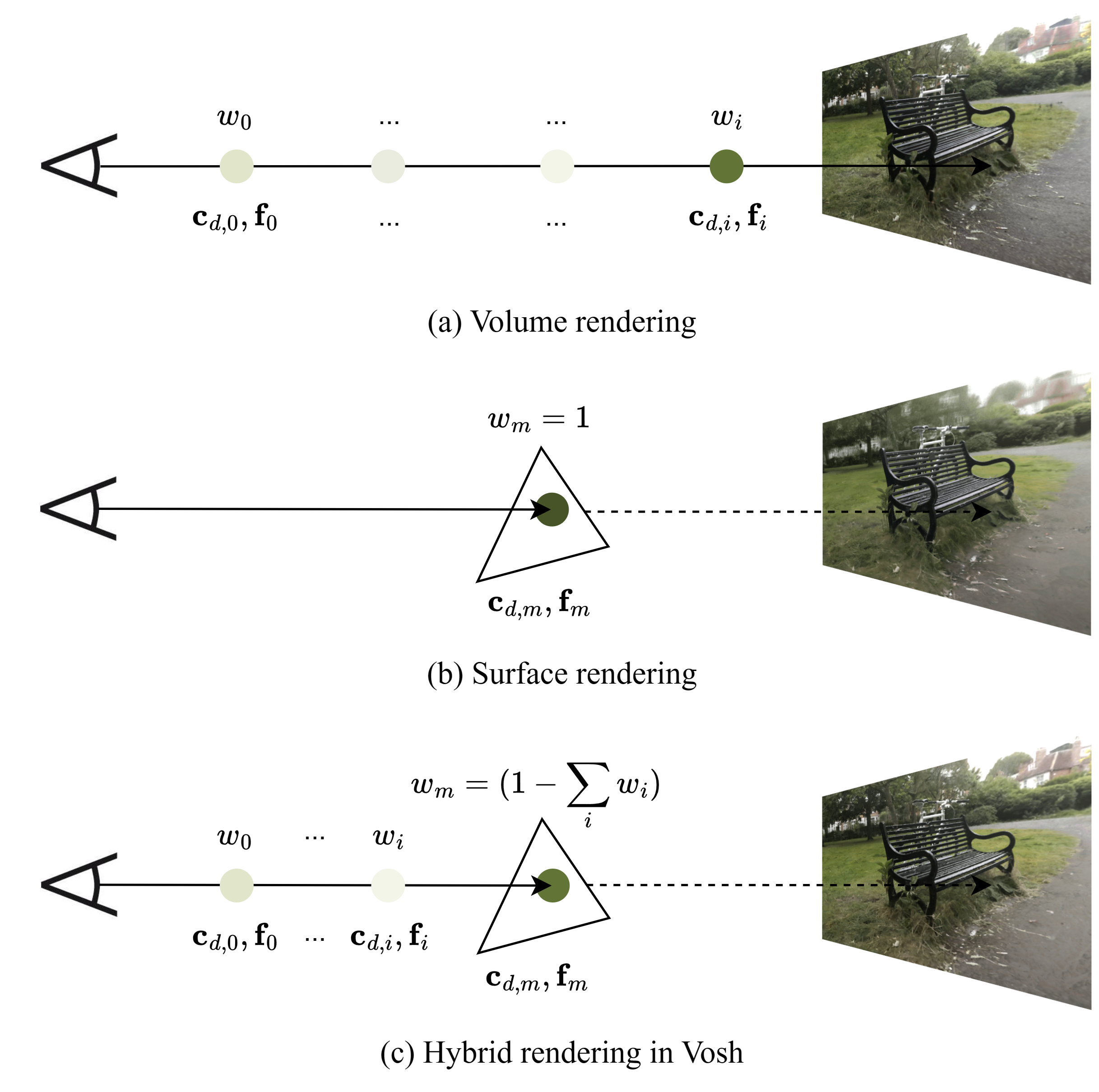}
  \caption{The proposed hybrid rendering (c) integrates volume rendering (a) and surface rendering (b) for the Vosh.}
  \label{fig:hybrid rendering}
\end{figure}

\renewcommand\arraystretch{1.5}
\begin{table*}[t]
\centering
\caption{Quantitative evaluation of the proposed method and SOTA methods on 3 outdoor scenes and 4 indoor scenes from Mip-NeRF 360 dataset~\cite{mipnerf360}. The methods based on hybrid representation are listed separately below. $^*$ means the method is not supported on the tested device.}
\label{tab:quality}
\begin{tabular}{c|ccc|ccc|cc}
 &
  \multicolumn{3}{c|}{Outdoor} &
  \multicolumn{3}{c|}{Indoor} &
  Intel Iris Xe 80EU &
  iPhone 14 Pro \\
\multirow{-2}{*}{Methods} &
  PSNR ↑ &
  SSIM ↑ &
  LPIPS ↓ &
  PSNR ↑ &
  SSIM ↑ &
  LPIPS ↓ &
  FPS @ 800$\times$600&
  FPS @ 347$\times$600\\ \hline
NeRF &
  21.46 &
  0.458 &
  0.515 &
  26.84 &
  0.790 &
  0.370 &
  -$^*$ &
  -$^*$ \\
Mip-NeRF 360 &
  25.92 &
  0.747 &
  0.244 &
  31.72 &
  0.917 &
  0.179 &
  -$^*$ &
  -$^*$ \\
Instant-NGP &
  23.90 &
  0.648 &
  0.369 &
  29.47 &
  0.877 &
  0.273 &
  -$^*$ &
  -$^*$ \\
Zip-NeRF &
  \textbf{26.87} &
  \textbf{0.807} &
  0.176 &
  \textbf{32.03} &
  \textbf{0.925} &
  \textbf{0.167} &
  -$^*$ &
  -$^*$ \\
3DGS &
  26.40 &
  0.805 &
  \textbf{0.173} &
  30.41 &
  0.920 &
  0.190 &
  -$^*$ &
  -$^*$ \\
SNeRG++ &
  24.48 &
  0.646 &
  0.358 &
  27.90 &
  0.837 &
  0.238 &
  22.0 &
  -$^*$ \\
MERF &
  24.41 &
  0.678 &
  0.298 &
  27.80 &
  0.855 &
  0.271 &
  17.3 &
  25.3 \\
BakedSDF &
  23.40 &
  0.577 &
  0.351 &
  27.20 &
  0.845 &
  0.300 &
  21.0 &
  -$^*$ \\
MobileNeRF &
  22.90 &
  0.524 &
  0.463 &
  25.75 &
  0.757 &
  0.453 &
  59.3 &
  153.3 \\
NeRF2Mesh &
  22.74 &
  0.497 &
  0.463 &
  25.20 &
  0.721 &
  0.340 &
  \textbf{138.7 }&
  \textbf{210.0} \\ \hline
\rowcolor[HTML]{FFFFFF} 
Adaptive Shells &
  23.17 &
  0.606 &
  0.389 &
  \textbf{29.19} &
  \textbf{0.872} &
  \textbf{0.285} &
  -$^*$ &
  -$^*$ \\
\rowcolor[HTML]{FFFFFF} 
VMesh &
  - &
  - &
  - &
  - &
  - &
  - &
  - &
  - \\
\rowcolor[HTML]{FFFFFF} 
Vosh-Light &
  23.73 &
  0.601 &
  0.378 &
  27.20 &
  0.802 &
  0.275 &
  \textbf{78.0} &
  \textbf{78.7} \\
\rowcolor[HTML]{FFFFFF} 
Vosh-Base &
  \textbf{24.18} &
  \textbf{0.623} &
  \textbf{0.364} &
  27.61 &
  0.818 &
  0.261 &
  38.0 &
  34.0
\end{tabular}
\end{table*}

\subsection{Optimization}

After meshing a portion of density field based on rending quality, we obtain an explicit surface suitable for representing regions in the scene with simple geometry and textures, as well as a optimized high-resolution voxel grid. By combining these two components, we can obtain a coarse hybrid representation with a significant amount of redundancy between the components. This may result in low rendering efficiency and potential rendering artifacts. So we further perform hybrid rendering optimization and voxel pruning based on the components for a more compact and efficient representation.

Compared to volume rendering, surface rendering only considers the intersection points between rays and surface, significantly reducing the time cost associated with ray marching. This inspires us to confine the ray marching in volume rendering to occur only before the rays hit the surface. Simultaneously, performing volume rendering between the viewpoint and the surface also provides possibilities for enhancing surface rendering quality. Therefore, we straightforwardly combine volume rendering and surface rendering, treating the surface points obtained through rasterization as the final points reached by ray marching in volume rendering. The hybrid rendering can be expressed by the following three equations:
\begin{equation}
    \label{eqn:4}
    \mathrm{\mathbf{C}}_{vosh}=\mathrm{\mathbf{C}}_{vosh,d}+\text{MLP}(\mathrm{\mathbf{C}}_{vosh,d}, \mathrm{\mathbf{F}_{vosh}, \mathrm{\mathbf{d}}}),
\end{equation}

\begin{equation}
    \label{eqn:5}
    \mathrm{\mathbf{F}}_{vosh}=\sum_iw_i\mathrm{\mathbf{f}}+w_m\mathrm{\mathbf{f}}_m,
\end{equation}

\begin{equation}
    \label{eqn:6}
    \mathrm{\mathbf{C}}_{vosh,d}=\sum_iw_i\mathrm{\mathbf{c}}_{d,i}+w_m\mathrm{\mathbf{c}}_{d,m},
\end{equation}
where $\mathrm{\mathbf{C}}_{vosh}$, $\mathrm{\mathbf{C}}_{vosh,d}$ and $\mathrm{\mathbf{F}}_{vosh}$ represent the color, diffuse color, and view-dependent features after hybrid rendering, respectively. $w_m=(1-\sum_iw_i)$ represents the rendering weight for the mesh, which is the remaining portion of the cumulative voxel weights. $\mathrm{\mathbf{c}}_{d,m}$ and $\mathrm{\mathbf{f}}_{m}$ represent the diffuse color and view-dependent features at the intersection point where the ray hits the surface. Three different rendering procedures are shown in Figure~\ref{fig:hybrid rendering}.

Based on the optimized mesh, we attempt to dynamically prune the voxel grid, forming a high-quality hybrid representation. Since the trained  mesh already possesses the capability to represent regions with simple geometry and textures, we need to consider how to simplify the voxel representation for the entire scene. We focus on retaining voxels in regions with complex geometry and textures to better complement the rendering of the hybrid representation with the mesh. Inspired by PlenOctrees~\cite{plenoctrees}, we propose a voxel pruning loss based on surface mesh weights. Specifically, we calculate the rendering weight of mesh $w_m$ for rays hitting the surface and force it close to 1. The implementation is as follows:
\begin{equation}
    \label{eqn:7}
    \mathcal{L}_{voxel}=\lambda_{voxel}\frac{1}{K}\sum_k(1-\text{exp}(w_m-1)),
\end{equation}
where $\lambda_{voxel}$ represents the weight of the voxel pruning loss, and $K$ is the number of samples taken before the rays hits the surface. 

The number of retained voxels is an important factor in determining the balance between rendering quality and speed. Concretely, retaining more voxels can enhance rendering quality but slow down rendering speed, while retaining fewer voxels can accelerate rendering speed but compromise rendering quality. Setting different voxel pruning weights to limit the volumetric density during ray marching is one of the methods to achieve different trade-offs between speed and quality. Additionally, voxels and mesh may overlap at the same spatial location, potentially affecting rendering quality and the volume of exported assets. To address this issue, we first calculate the occupancy grid of the meshes under the training poses and filter the voxels in the mesh-occupied grid out during rendering and baking.
We also adjust the ratio of sparse voxels in the hybrid representation by setting different resolutions $r_{mesh}$ for the mesh-occupancy grid, which can also help suppress density of voxel component in Vosh.

\section{Real-Time Rendering}
The real-time rendering based on Vosh is realized as a JavaScript 3D web application, leveraging the capabilities of the MERF real-time renderer. The rendering process is also hybrid and structured into two passes: the rasterization pass and the ray marching pass. Both passes are employed to make the rendering result consistent between training on the GPU and inference on the web.

In the rasterization pass, meshes are uniformly shaded. The fragment shader takes charge of sampling appearance textures, incorporating diffuse color and view-dependent feature. The output of the rasterization pass includes corresponding diffuse, view-dependent features and depth map. This output is then used in the subsequent ray marching pass.

In the ray marching pass, we also employ a multi-resolution hierarchical structure for the occupancy grid introduced in SNeRG~\cite{snerg} to mark the position of occupied voxels. This structure creates binary mask images in multiple resolution levels through max-pooling. We use the multi-resolution occupancy grid to perform spatial skips during ray marching. Simultaneously, the depth map obtained in the rasterization pass further reduces the sampling distance.

For each spatial coordinate after ray marching, we compare the transformed depth with the depth map from the rasterization pass to determine whether the ray has intersected the surface position. If the ray reaches the surface of the mesh, the marching stops, and the accumulated result of ray marching is weightedly combined with the appearance result from the rasterization pass. Leveraging the surface depth information from the rasterization pass proves instrumental in significantly diminishing the number of required sampling points during ray marching, thereby effectively reducing the computational cost associated with volume rendering.

\begin{figure*}[tbp]
  \includegraphics[width=\textwidth]{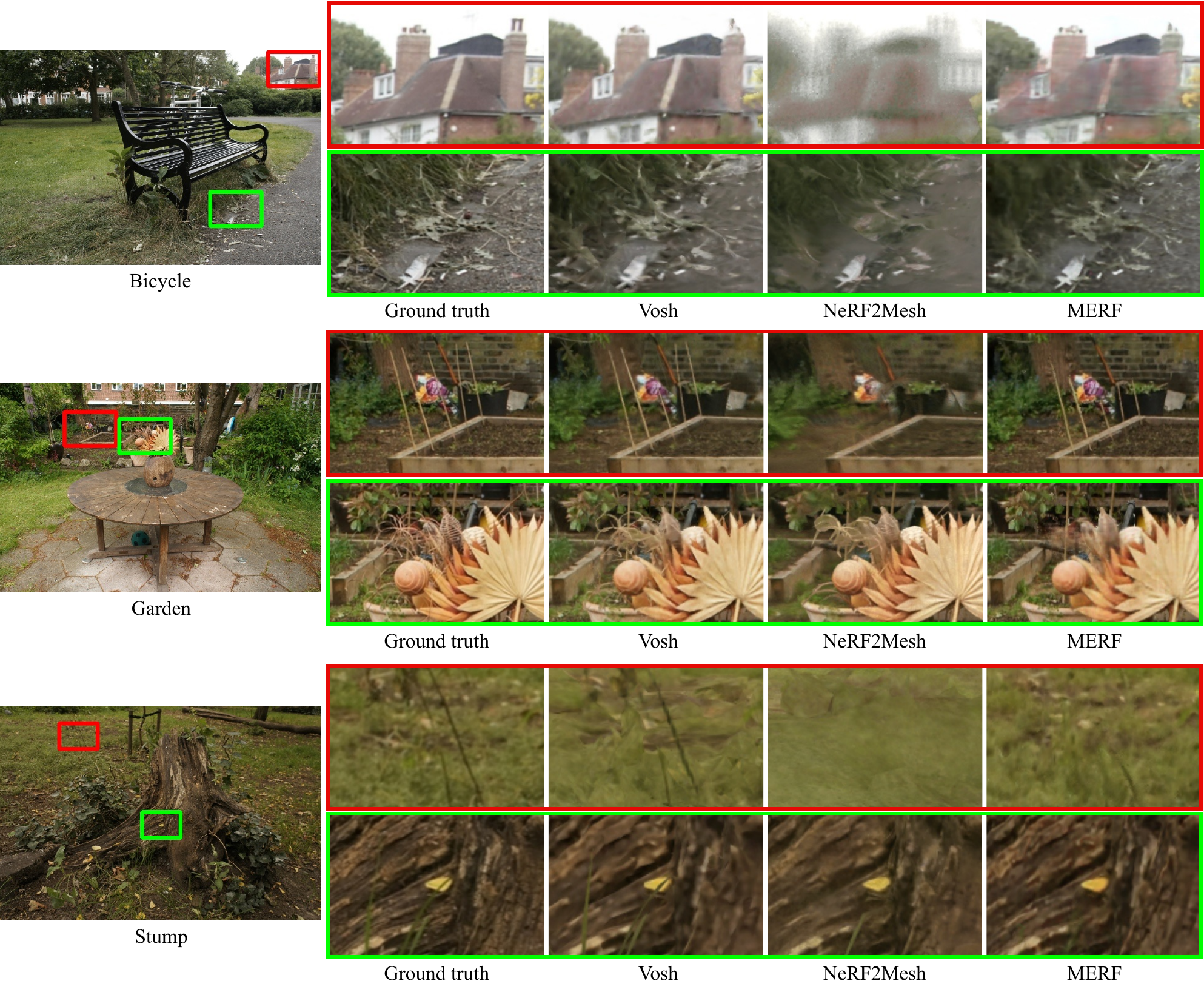}
  \caption{The rendering results and zoomed-in images in outdoor scenes obtained by our method, as well as some SOTA methods based on voxels or mesh representation.}
  \label{fig:quality}
\end{figure*}

\section{Experiments}

We have implemented real-time view synthesis using the proposed hybrid representation Vosh, and carried out comprehensive evaluations encompassing both rendering quality and speed. We also compare our method with state-of-the-art (SOTA) methods, like the classical offline methods (NeRF~\cite{nerf}, Mip-NeRF 360~\cite{mipnerf360}, Instant-NGP~\cite{instantngp}) and real-time methods (MobileNerf \cite{mobilenerf}, NeRF2Mesh~\cite{nerf2mesh}, BakedSDF~\cite{bakedsdf}, MERF~\cite{merf}, SNeRG++~\cite{snerg}). 
As done by SOTA methods, we use the challenging unbounded dataset Mip-NeRF 360~\cite{mipnerf360} for evaluation. 
Here, we specifically list the two most relevant hybrid representation based methods of Adaptive Shells~\cite{adaptiveshells} and VMesh~\cite{vmesh} as shown in Table~\ref{tab:quality}. For Adaptive Shells, we obtain quantitative results from its paper, while VMesh mostly handles individual objects or bounded scenes, while our method focuses more on real scenes. 
Next, we elaborate the details of the experiments.

\begin{figure*}[tbp]
  \includegraphics[width=\textwidth]{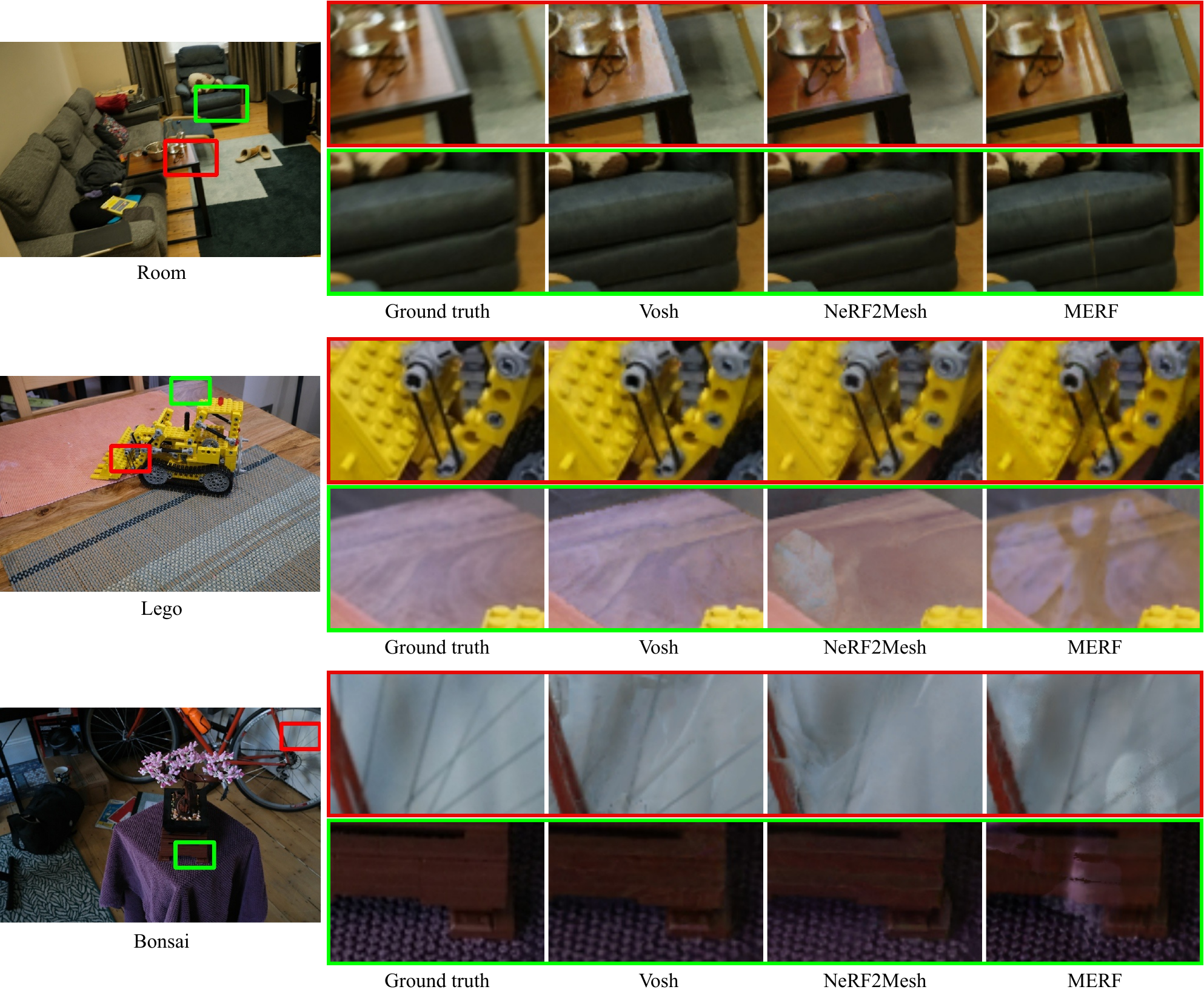}
  \caption{More rendering results and zoomed-in images in indoor scenes obtained by our method, as well as some SOTA methods based on voxels or mesh representation.}
  \label{fig:quality_indoor}
\end{figure*}

\subsection{Implementation details}
In the stage of training initial voxels, we use the classic MSE loss to measure pixel-level color differences. We use the same optimizer settings as MERF, e.g., 25000 steps for optimization and initially setting learning rates to 1e-2, which warmup on the first 100 steps and ultimately decay to 1e-3. We use 1024 as the resolution of the marching cubes and clean mesh with similar operations in NeRF2Mesh. 

In the stage of meshing density field, we use 20000 steps for mesh refinement. For the process of meshing density field, we set $p_{remove}=0.9$ for Vosh-Base and $p_{remove}=1.0$ for Vosh-Light. After the meshing stage, some messy floating faces may be obtained. We use similar parameters to repeat the operation of cleaning the mesh in the previous stage. And we encourage voxels to fill possible holes in the mesh by using L2 regularization terms with rendering weights for background colors.

There are 20000 steps for Vosh optimization. For the Vosh-Base model, we set voxel pruning loss $\lambda_{voxel}=0.001$ and disable the binary mesh-occupancy grid. For the Vosh-Light model, we use voxel pruning loss $\lambda_{voxel}=0.01$ and the resolution $r_{mesh}=128$ for mesh occupancy grid as hyper-parameters.

Besides, we did all experiments on a single NVIDIA RTX 4090. Three training stages take approximately 66, 35 and 58 minutes respectively. During the training process, the peak video memory usage for optimizing trainable parameters is about 11 Gigabytes. It should be noted though the training phase requires high-performance hardware, the inference phase can be performed directly on mobile devices with limited resources.

\subsection{Rendering quality comparison}

We conduct both quantitative and qualitative evaluation of rendering quality for view synthesis obtained by the proposed method and SOTA methods. Quantitative evaluation uses the metrics of PSNR, SSIM~\cite{ssim} and LPIPS~\cite{lpips} for the assessment of rendering quality. Table~\ref{tab:quality} shows the statistics of the quantitative evaluation of different methods.

Vosh demonstrates superior rendering quality in real-time rendering on mobile devices, even surpassing some methods that cannot achieve real-time rendering on mobile devices. For the outdoor scenes, Vosh enhances rendering quality over mesh-based methods by leveraging sparse voxels through the use of a contraction function (see Figure~\ref{fig:quality} and \ref{fig:quality_indoor}). 
Rendering quality of Vosh in outdoor scenes also exceeds that of Adaptive Shells, which is also based on hybrid representation. The main reason is that Adaptive Shells constrain volume rendering to the regions near the surface, while in outdoor scenes, pure volume rendering is mostly of necessity. 
However, the limitations in representational capacity of the methods using tiny view-dependent MLPs limit rendering quality in indoor scenes.
While Vosh achieves real-time rendering on mobile devices by employing a tiny view-dependent MLP, it has slightly lower rendering quality in indoor scenes compared to MLP-based methods.

We provide a qualitative comparison with SOTA methods based on voxels~\cite{merf} and mesh~\cite{nerf2mesh} on the examples in Figure~\ref{fig:quality}. We select one distant view and one close-up view with local magnification for better exhibition. The mesh-based methods often lack the ability to represent background regions far from the camera location, whereas volume-based methods necessitate higher computational and storage costs to achieve enhanced rendering quality. Contrarily, our method improves rendering speed while essentially maintaining the rendering quality comparable to that of volume-based methods. We refer the reader to the companion video for dynamic demonstrations of the results. 

\subsection{Rendering speed comparison}

We evaluated the performance of the methods applicable for rendering on platform-independent mobile devices, including SNeRG++~\cite{snerg}, MERF~\cite{merf}, BakedSDF~\cite{bakedsdf}, MobileNeRF~\cite{mobilenerf}, and NeRF2Mesh~\cite{nerf2mesh}. We conduct testing on iPhone 14 Pro equipped with the A16 chip with a resolution of 347$\times$600 and on Lenovo Yoga 14s 2021, which features Intel Iris Xe Graphics 80EU (28W, integrated in Intel Core i5-11350H) and operates with a resolution of 800$\times$600. For SNeRG++ and MERF, we follow their configurations with progressive upsampling turned off. 

Rendering speed tests for all comparative methods are conducted on the Mip-Nerf 360 dataset~\cite{mipnerf360} in three outdoor scenes, using identical device settings, resolutions, and viewpoints. As shown on the right side of Table~\ref{tab:quality}, our method can run with the real-time frame rate (\textgreater 30 FPS) on both iPhone 14 Pro and Iris Xe 80EU. 
Adaptive Shell~\cite{adaptiveshells} can only achieve real-time rendering at 1080p resolution using an RTX 4090 GPU (36.18 FPS from its paper). Our method can achieve faster frame rates at the same resolution using an GTX 1660 Ti GPU.
The volume-base methods have slow rendering speeds due to the necessity to sample each alive voxel traversed by every ray. 
BakedSDF~\cite{bakedsdf} is difficult to render on mobile phones and laptops with limited VRAM due to its substantial assets. Specifically, BakedSDF has about 10 million triangles to represent unbounded scenes, mostly for complex backgrounds. Vosh, leveraging the efficiency of voxel component, reduces the need for triangles in low-quality and space-consuming background areas, using only about one-tenth as many. This greatly shortens the time of rasterization, providing an opportunity to enhance the performance based on sparse voxels. 
NeRF2Mesh~\cite{nerf2mesh} achieves rendering speeds that reach the maximum refresh rates of their respective devices (60Hz for Safari on iPhone 14 Pro and 90Hz for Lenovo Yoga 14s 2021). We make statistics on their potential rendering speeds using overlapping multiple passes.

\subsection{Comparison with VMesh}
\begin{figure}[tbp]
  \includegraphics[width=\linewidth]{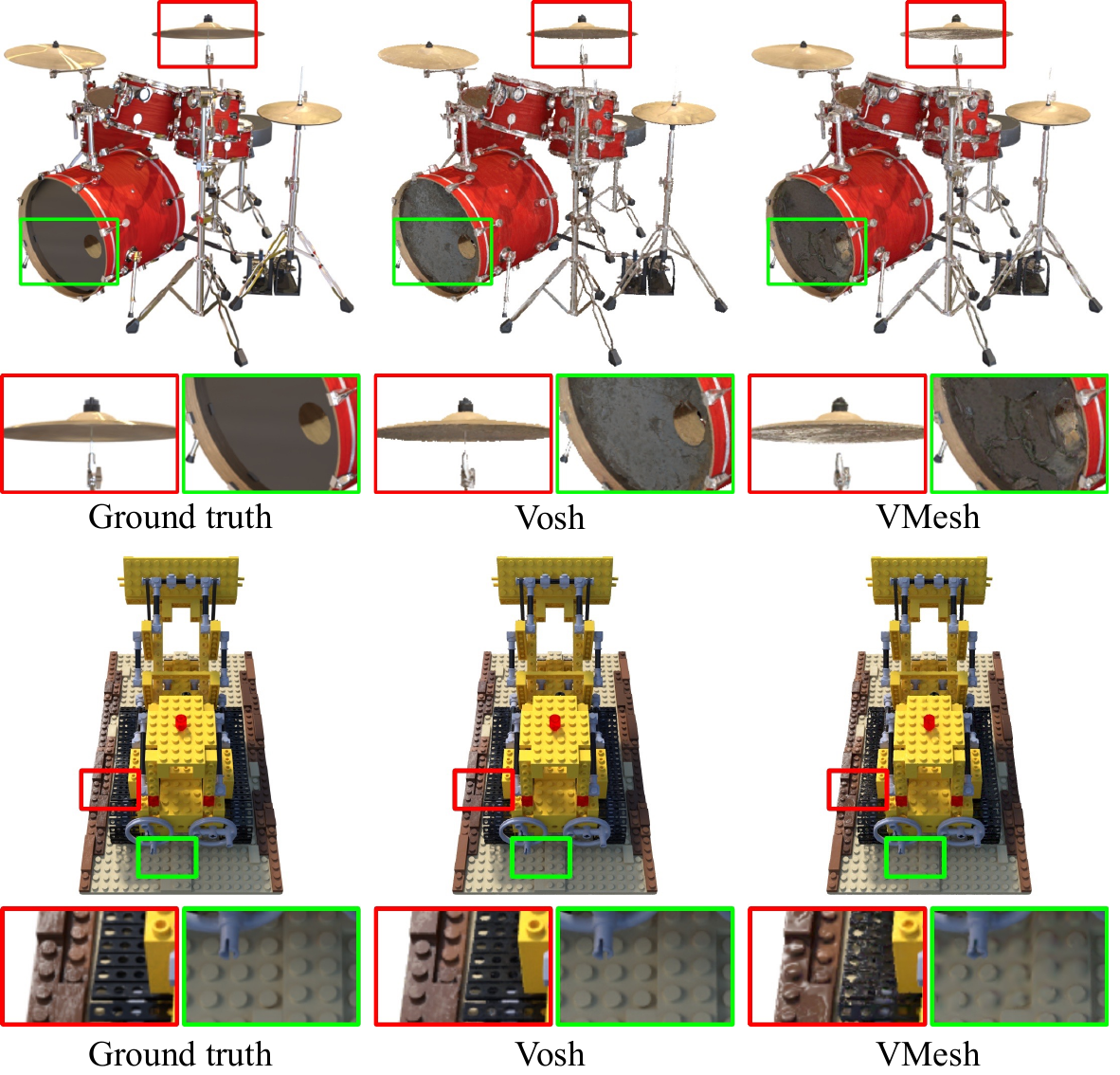}
  \caption{The rendering results and zoomed-in images on the NeRF-Synthetic dataset~\cite{nerf} from our method and VMesh~\cite{vmesh}.}
  \label{fig:vmesh cmp}
\end{figure}

It is worth noting that VMesh~\cite{vmesh} is also built on the hybrid representation of volume and mesh, but primarily designed for rendering individual objects or scenes with boundaries, posing challenges in handling unbounded scenes like the examples in Figure~\ref{fig:quality}. 
To highlight the effectiveness of the proposed method, we particularly compare Vosh with VMesh on NeRF-Synthetic dataset~\cite{nerf}. The speed test in NeRF-Synthetic is based on the Lenovo Yoga 14s 2021 equipped with Iris Xe Graphics 80EU at a resolution of 800$\times$600.
Due to the particularity of the NeRF-Synthetic dataset, the mesh extracted based on volume representation can usually achieve higher rendering quality than the volume representation itself after optimization~\cite{nerf2mesh}. So we mesh the density field of the entire voxel grid, which means that the $p_{remove}$ is fixed to $1$.
Our method has better performance than VMesh in the aspect of rendering quality or speed, as shown in Table~\ref{tab:vmesh_synthetic} for quantitative comparison and Figure~\ref{fig:vmesh cmp} for qualitative comparison.

\renewcommand\arraystretch{1.8}
\begin{table}[tbp]
\caption{Quantitative evaluation of the proposed method and VMesh~\cite{vmesh} in NeRF-Synthetic dataset~\cite{nerf}.}
\label{tab:vmesh_synthetic}
\centering
\begin{tabular}{l|lll|l}
Methods    & PSNR  & SSIM  & LPIPS  & FPS \\ \hline
VMesh      & 30.75 & 0.948 & 0.059  & 261.0 \\
Vosh-Base  & 30.96 & 0.948 & 0.064  & 85.0 \\
Vosh-Light & 30.77 & 0.947 & 0.065  & 281.8\\
\end{tabular}
\end{table}

\begin{figure}[tbp]
  \includegraphics[width=\linewidth]{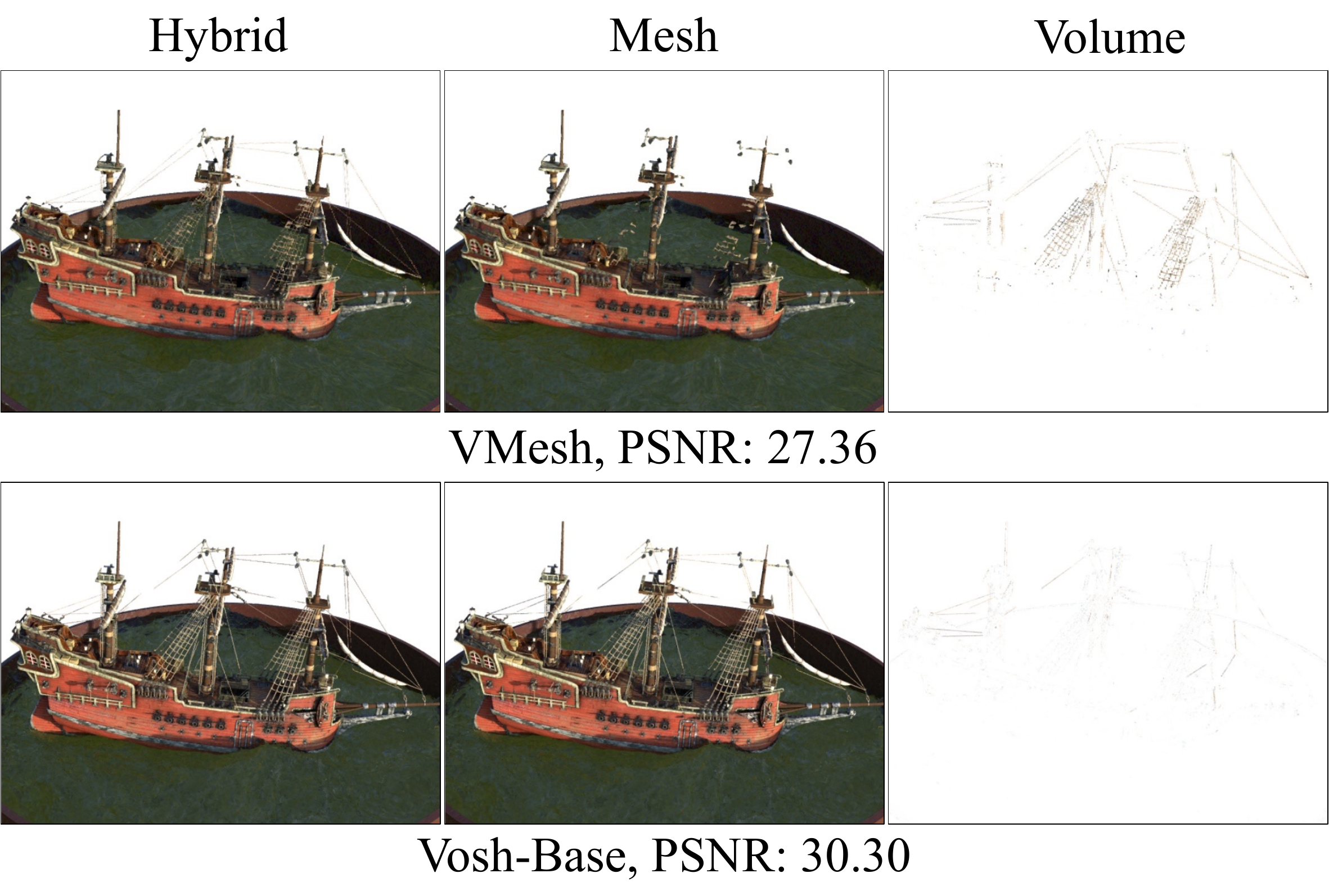}
  \caption{The hybrid, mesh based and volume based rendering results obtained by our method and VMesh~\cite{vmesh} on the NeRF-Synthetic dataset~\cite{nerf}. The PSNR in the figure represents the evaluation result of the current images.}
  \label{fig:vmesh cmp2}
\end{figure}

We also provide the rendering results for the mesh part and the volume part separately, in addition to the rendering results of the hybrid representation of VMesh and Vosh, in Figure~\ref{fig:vmesh cmp2}. 
Our reconstruction method of hybrid representation based on volumetric density field can restore more comprehensive scene information, thereby utilizing the extracted mesh to represent thin structures that are difficult to reconstruct in SDF, such as fishing nets and ropes in the figure. Hence, our method usually gains better rendering quality than VMesh.

\subsection{Storage comparison}
We conduct a statistical analysis of the graphics memory consumption and disk usage of the SOTA methods and the proposed method on the outdoor scenes of Mip-NeRF 360~\cite{mipnerf360}, as shown in Table~\ref{tab:storage}. We use the occupying grid based approach proposed by SNeRG~\cite{snerg} to store voxels, and use methods similar to MobileNeRF~\cite{mobilenerf} and NeRF2Mesh~\cite{nerf2mesh} to store mesh assets. 
Since we do not use the triplane representation which is the core contribution of MERF~\cite{merf}, but only recover the scene information and generate a hybrid representation based on the classic voxel representation SNeRG~\cite{snerg}, it appears to be slightly larger than MERF in terms of disk storage.
Although our method does not make any additional optimizations in terms of memory, it still has certain advantages compared to methods with similar rendering quality or speed.

\renewcommand\arraystretch{1.8}
\begin{table}[tbp]
\centering
\caption{Memory consumption and disk usage of the SOTA methods in outdoor scenes on the Mip-NeRF 360 dataset.}
\label{tab:storage}
\begin{tabular}{c|c|c}
Methods     & VRAM (MB) ↓ & Disk (MB) ↓ \\ \hline
SNeRG++     & \textbf{340} & 624         \\ \hline
MERF        & 360         & 149         \\ \hline
NeRF2Mesh   & 426         & \textbf{123}         \\ \hline
MobileNeRF  & 927         & 345         \\ \hline
BakedSDF    & 953         & 807         \\ \hline
3DGS        & 3701        & 1270        \\ \hline
Vosh-Light  & 490      & 232         \\ \hline
Vosh-Base   & 506     & 488        
\end{tabular}
\end{table}

\subsection{Ablation study}
We conduct ablation experiments on the meshing density field and voxel pruning parts of the hybrid representation on outdoor scenes of Mip-NeRF dataset~\cite{mipnerf360}, as shown in Table~\ref{tab:ablation_normal} and Table~\ref{tab:ablation}. The assessment involves a quantitative comparison, considering both rendering quality and speed.

We compare the variants of different optimization methods for differentiable surface optimization in Table~\ref{tab:ablation_normal}. In addition to rendering quality, we also report the number of faces (in millions) of different variants . It can be seen that optimizing the geometry of the mesh based on differentiable rendering can significantly improve rendering quality. 
The experiments on surface operations indicate that it can better preserve rendering quality without mesh simplification, but might result in a large number of triangles. It might also cause some loss of rendering quality without performing mesh subdivision. So we use mesh subdivision and simplification based on rendering quality and normal angle change to significantly improve rendering quality while decreasing the number of triangles.

\begin{figure}[tbp]
  \includegraphics[width=\linewidth]{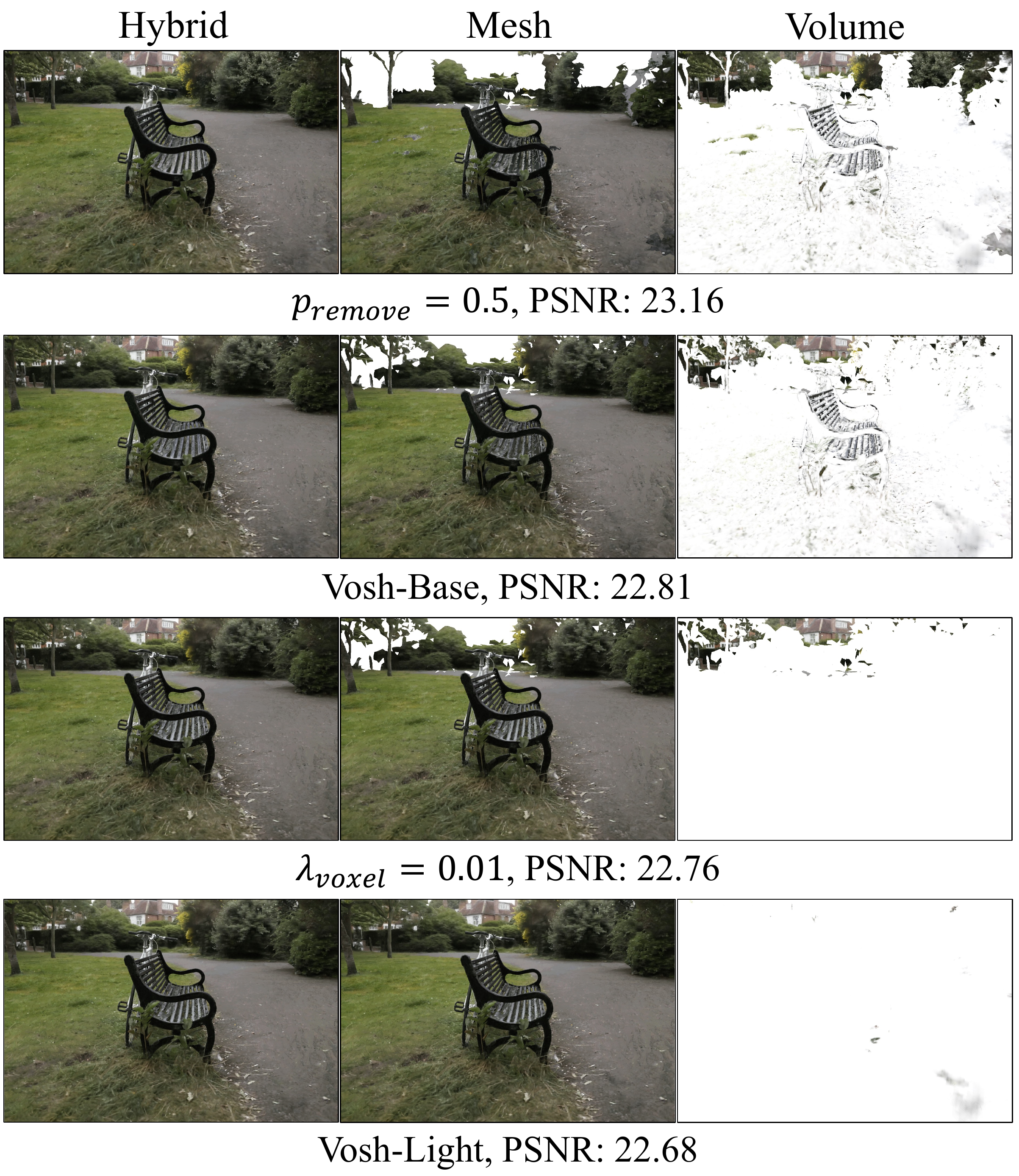}
  \caption{The hybrid, mesh based and volume based rendering results obtained by our method on Mip-NeRF 360 dataset~\cite{mipnerf360}. The PSNR in the figure represents the evaluation result of the current images.}
  \label{fig:abalation}
\end{figure}

Compared with the mesh optimization in NeRF2Mesh~\cite{nerf2mesh}, our method additionally considers the angle of normals between adjacent triangles during the optimization, which can improve rendering quality while reducing the number of triangles. At the same time, our meshing strategy based on rendering quality at the same stage also helps generating the high quality hybrid representation as shown in Fig~\ref{fig:abalation}.

\begin{table}[tbp]
\caption{The ablation study in the differentiable surface refinement.}
\label{tab:ablation_normal}
\centering
\begin{tabular}{ll|lll|l}
                      &     Variants     & PSNR  & SSIM  & LPIPS  & \# faces \\ \hline
\multirow{5}{*}{\rotatebox{90}{Mesh}}
& Baseline & 22.04 & 0.475 & 0.448 & 1.76 \\  \cline{2-6}
& Refine w/o subd & 23.52& 0.597& 0.379&1.44\\  \cline{2-6}
& Refine w/o deci & 23.61& 0.601& 0.377& 2.14\\ \cline{2-6}
& Refine w/o norm & 23.54& 0.594& 0.381&1.83\\ \cline{2-6}
& Refine & 23.57& 0.598& 0.379& 1.69\\
\end{tabular}
\end{table}

\begin{table}[tbp]
\caption{The ablation study in the construction of hybrid representation.}
\label{tab:ablation}
\centering
\begin{tabular}{l|l|lll|ll}
                               & Variants               & PSNR  & SSIM  & LPIPS & \# faces & \# voxels \\ \hline
\multirow{4}{*}{\rotatebox{90}{Meshing}}       & $p_{remove}=0.5$       & \textbf{24.49} & 0.642 & \textbf{0.357} & 0.81     & 98999     \\
                               & $p_{remove}=1.0$       & 23.95 & 0.600 & 0.384 & 1.69     & 47631  \\ 
                               & Refined mesh      & 23.57 & 0.598 & 0.379 & 1.69     & 0 \\ 
                               & Voxel grid          & 24.48 & \textbf{0.646} & 0.358 & 0     & 126533 \\ \hline
\multirow{4}{*}{\rotatebox{90}{Voxel prune.}} & $\lambda_{voxel}=0$    & \textbf{24.23} & \textbf{0.625} & \textbf{0.358} & 1.41     & 74317     \\
                               & $\lambda_{voxel}=0.01$ & 23.90 & 0.606 & 0.377 & 1.41     & \textbf{36255}     \\
                               & $r_{mesh}=256$           & 23.80 & 0.604 & 0.377 & 1.41   & 41743     \\
                               & $r_{mesh}=512$           & 23.94 & 0.609 & 0.373 & 1.41   & 52280  \\ \hline   
                               & Vosh-Base              & 24.18 & 0.623 & 0.364 & 1.41     & 57438     \\ 
                               & Vosh-Light             & 23.74 & 0.602 & 0.377 & 1.41     & 2489
\end{tabular}
\end{table}

To better illustrate the effectiveness of each component in the construction process of our hybrid representation, we compare Vosh-Base with its variants in Table~\ref{tab:ablation}, which includes the threshold $p_{remove}$ in meshing based on rendering quality, the weight of the voxel pruning loss $\lambda_{voxel}$, and the resolution $r_{mesh}$ for mesh occupancy grid. 
We also report the number of faces (in millions) that make up the mesh component and the number of alive voxels that make up the voxels component in the Vosh representation. 

In the section of meshing density field, we compare the variants which $p_{remove}=0.5$ and $1.0$ with the Vosh-Base model with $p_{remove}=0.9$. Meanwhile, in addition to the rendering results of the initial voxel grid, we also provide refined rendering results for all surfaces obtained by meshing the entire voxel grid. When setting $p_{remove}=1.0$, it means that the entire density field is meshed, not based on rendering quality. The results show that this can only achieve a small improvement in quality. When setting $p_{remove}=0.5$, our hybrid representation that combines the advantages of voxel and mesh components achieve higher or comparable quality rendering results than the initial voxel grid.

In the section of voxel pruning, we compare the variants $\lambda_{voxel}=0$ and $0.01$ with the Vosh-Base model with $\lambda_{voxel}=0.001$. We also list the rendering results of the variants $r_{mesh}=256$ and $512$, while Vosh-Base model turns off the check of mesh occupancy grid. Experimental results suggest that increasing the weight of voxel pruning loss or reducing the resolution of the mesh occupancy grid aids in compressing the number of alive voxels, leading to speed benefits at the expense of rendering quality.

At the same time, we also provide the hybrid, mesh based and volume based rendering results of several representative variants of Vosh, as shown in Figure~\ref{fig:abalation}. 
We use Vosh-Base in the second line as the benchmark, which sets $p_{remove}$ to 0.9 and $\lambda_{voxel}$ to 0.001. As a comparison, the first line only changes $p_{remove}$ to 0.5 and the third line only changes $\lambda_{voxel}$ to 0.01. On the basis of Vosh-Base, Vosh-Light sets $\lambda_{voxel}$ to 0.01, $p_{remove}$ to 1.0 and $r_{mesh}$ to 128. 
Based on the qualitative comparison in this figure, it can be seen that we can extract surfaces with higher reconstruction quality by determining the region where the density field meshes based on the difference in rendering quality through the threshold $p_{remove}$. And the balance between quality and speed in the hybrid representation model can be adjusted by setting the different weight $\lambda_{voxel}$ of the voxel pruning loss and the resolution $r_{mesh}$ for mesh occupancy grid.

\subsection{Limitations}
Although the proposed hybrid representation Vosh brings valuable advantages to real-time view synthesis, it is not without limitations. We adopt the same view-dependent MLP as SNeRG and MERF, consequently sharing some similarities with their drawbacks. Notably, it faces challenges such as the incapability to model view-dependent colors for translucent objects, and represent extremely large scenes or objects with complex reflections. These limitations, in turn, may result in potential degradation of mesh optimization quality, thereby impacting the rendering quality based on the hybrid representation.

\section{Conclusion}
We have presented a hybrid representation Vosh, composed of both voxels and mesh, for NeRF-based real-time view synthesis. By capitalizing on the strengths of voxel and mesh components in our representation, we attain a controllable balance between rendering quality and speed. Experimental evidence demonstrates that our approach outperforms state-of-the-art methods, showcasing exceptional flexibility for deployment across a range of mobile devices.

In our future work, we intend to explore more efficient data structures for organizing and storing the hybrid representation, aiming to optimize memory consumption and further enhance rendering efficiency. Additionally, we are considering more compact integration of classic representations to achieve fast and high-quality novel view synthesis specifically tailored for mobile devices.

\ifCLASSOPTIONcaptionsoff
  \newpage
\fi

\bibliographystyle{IEEEtran}

\bibliography{reference}

\vfill

\end{document}